\documentclass[10pt,twocolumn,letterpaper]{article}

\usepackage{cvpr}
\usepackage{times}
\usepackage{epsfig}
\usepackage{graphicx}
\usepackage{amsmath}
\usepackage{amssymb}

\usepackage{graphicx} 

\DeclareMathOperator*{\argmin}{argmin}   
\DeclareMathOperator*{\argmax}{argmax}   

\usepackage{amsfonts}
\usepackage{multirow}
\usepackage{hyperref}
\usepackage{wasysym}
\usepackage{mathtools}

\usepackage{amssymb}
\usepackage{pifont}
\newcommand{\cmark}{\ding{51}}%
\newcommand{\xmark}{\ding{55}}%

\usepackage{xcolor}

\newcommand{\cmmnt}[1]{}

\cvprfinalcopy 

\def\cvprPaperID{****} 

\begin{document}

\title{A Survey on Assessing the Generalization Envelope of Deep Neural Networks: Predictive Uncertainty, Out-of-distribution and Adversarial Samples}

\author{Julia Lust and Alexandru P. Condurache\\
Robert Bosch GmbH, Automated Driving Research\\
University of L\"ubeck, Institute of Signal Processing\\
{\tt\small JuliaRebecca.Lust, AlexandruPaul.Condurache@de.bosch.com}
}

\maketitle

\begin{abstract}
Deep Neural Networks (DNNs) achieve state-of-the-art performance on numerous applications. However, it is difficult to tell beforehand if a DNN receiving an input will deliver the correct output since their decision criteria are usually nontransparent. A DNN delivers the correct output if the input is within the area enclosed by its generalization envelope. In this case, the information contained in the input sample is processed reasonably by the network. It is of large practical importance to assess at inference time if a DNN generalizes correctly. Currently, the approaches to achieve this goal are investigated in different problem set-ups rather independently from one another, leading to three main research and literature fields: predictive uncertainty, out-of-distribution detection and adversarial example detection. This survey connects the three fields within the larger framework of investigating the generalization performance of machine learning methods and in particular DNNs. We underline the common ground, point at the most promising approaches and give a structured overview of the methods that provide at inference time means to establish if the current input is within the generalization envelope of a DNN.
\end{abstract}

\section{Introduction}
\begin{figure}[t]
 	\centering
 	\includegraphics[width=0.47
 	\textwidth]{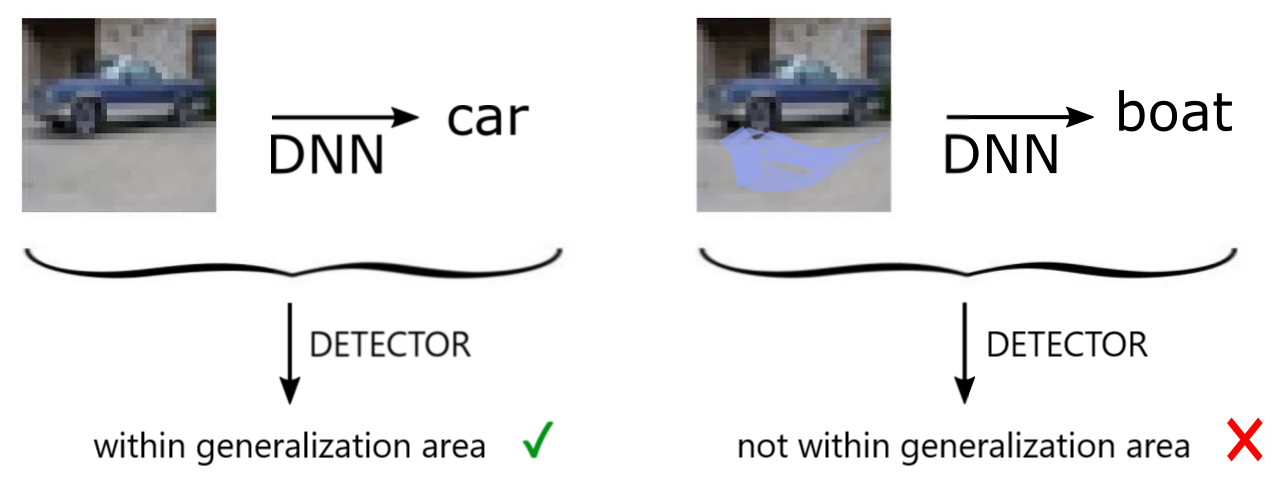}
 	\caption{A detector deciding at inference time for a given data sample if it is within the generalization area of a DNN.}
 	\label{Fig:detector}
 \end{figure}

%
%
%
%
Generalization is the ability of a classifier to correctly predict the class of previous unseen data points. In machine learning the classifier's decisions are derived from a training set. How well the derived decision criteria generalize is tested by using an additional test set containing unseen data, which is sampled from the same distribution as the training set. A good performance on the training set in combination with a poor performance on the test set is usually related to a lack of generalization ability. After the development of a classifier is finished, a poor performance on a data sample at inference time can often be traced back to a distribution shift between the development and inference data. For improved classification performance it is important to detect such situations in which the input has left the generalization area of the classifier.\\
Statistical analysis of the generalization potential of a classifier for the purpose of optimizing its parameters during training, has led to well known results such as the Probably Approximately Correct (PAC) Theory \cite{valiant1984theory}. Generalization was investigated as a combination of properly sampled training data and simpler decision methods.\\
Recently, Deep Neural Networks (DNNs) are the most successful machine learning methods for various tasks such as computer vision, speech recognition and object-detection \cite{deng2014deep}, \cite{lecun2015deep}, \cite{girshick2015fast}. DNNs make use of non-linearities in combination with simple matrix multiplications and an efficient data driven training procedure. This allows for complex deep structures and hence grasping the connection between the input and the output is usually not trivial. The deeper and hence more complex the neural network is, the less transparent and comprehensible is its correlation-based behavior. Usually humans, using their mostly causality-based intuition, are not able to tell why a DNN produced a certain output, which would be a major step in understanding their generalization behavior.\\
A possible solution to this dilemma is to decide for each input at inference time, if the DNN is likely to predict the correct output. Consequently research currently focuses on developing dedicated, separated generalization-detector methods that detect if the current input is within the corresponding generalization area of DNNs. In general this is done by analyzing the relationship between the input sample and the training set eventually complemented by an analysis of the information flow within the Network. These approaches largely replaced methods where a confidence in the result is computed from the DNN themselves (e.g., with the help of a softmax layer), as these tend to be overconfident \cite{goodfellow2014explaining}, \cite{nguyen2015deep}.\\
A generalization detector is illustrated in Figure \ref{Fig:detector}. A DNN receives an input on which it performs its classification task. The generalization detector now has to decide, based on the input and the behavior of the DNN, if this input is within the generalization area.
The intention of this paper is to give an overview and compare methods that decide at inference time for a given data sample if it is within the generalization area of a DNN. Even though this setup is valid for any application of DNNs, we concentrate here on image classification, as currently one of the best investigated areas in that field.\\
The current literature in the field of generalization detection methods at inference time can be split into three main fields: predictive uncertainty, out-of-distribution detection and adversarial example detection. Predictive uncertainty mainly concentrates on set-ups involving at inference time data sampled from the same distribution as the training data. The goal is to assign a high uncertainty to the samples that lead to a misclassification. Out-of-distribution detection deals with data that is different to the training data in a principled manner. The goal here is to detect such data at inference time. The adversarial example detection field concentrates on detecting samples that are carefully generated in order to fool the DNN.\\
Until now those research areas were usually treated seperatly rather ignoring their common root cause of deficient generalization performance. Methods belonging to each group are only evaluated on its corresponding setup. We strive here for a complete overview of all different sectors and their corresponding methods on the grounds of their common generalization-related root cause.\\
 In Section \ref{Sec:Definition} we introduce relevant concepts and the mathematical definitions necessary to explain the detection methods surveyed in Section \ref{Sec:Methods}. The related work is described in Section \ref{Sec:Related}. We list different properties and discuss similarities and differences across the surveyed fields in Section \ref{Sec:Discussion}. Finally we draw our conclusion in Section \ref{Sec:Conclusion}.
\section{Preliminaries}
\label{Sec:Definition}
In this section, we first give an overview on generalization and we describe the setup of our survey focusing on concepts and mathematical notations.
\subsection{Generalization Overview}
\label{Sec:Generalization}
Consider a classifier that is trained on a training set sampled from an unknown distribution. We do not know a-priori if the training set represents a proper sample or not - where a proper sample is one that allows us to perfectly reconstruct the unknown distribution. We wish that the classifier that we devise using the potentially improper training sample is able to generalize, which means that its performance at inference time is good on any sample of the distribution.\\
One possibility to obtain good generalization performance is to use a stable classifier. A stable classifier has limited variability with respect to the training set. Thus similar classifiers are found by a convergent training procedure even when the training data is (very) different. Not being able to closely 'follow' the training set, this classifier will hardly overfit and at the same time, the chances that the performance observed on the training set is representative for the performance on all data are higher. From a certain point of view, such a classifier has a good generalization performance in the sense that there is a good chance that the error on all data can be estimated using only the training set, even if this error is large and thus its classification performance on any sample of the distribution is low. We propose to call this property of being able to estimate the true performance on the training set \textit{stability} to keep it separated from generalization as introduced earlier.\\ 
The usual way to obtain a stable classifier is to limit the hypothesis space that it covers. The parameters of the classifier are then found such as to minimize the empirical error computed on the training set. Clearly, this approximation is better the more (properly sampled) data is available. Thus, in this case generalization is understood mainly in relation to properties of the function implementing the classifier and also in  relation to the empirical error. In practice we need to control the modeling capacity of the function space to which the classifier belongs in relation to the cardinality of the training sample (assuming it is a proper sample), while minimizing the training error. Should we succeed and achieve minimal (ideally zero) training error, we can assume that we have found the correct solution to our classification problem, as the classifier does not use its modeling capacity to learn by heart and generalizes well in the sense that it has good classification performance on unseen data. This approach with its bias-variance trade-off flavor, inspired the support vector machines \cite{vapnik1999overview}, \cite{vapnik2013nature} whose linear decision surface is constructed to ensure high generalization ability \cite{cortes1995support}.\\
In the DNN context, the modeling capacity is infinite \cite{haykin2007neural}, but still we would want it to generalize, i.e., perform well on previously unobserved input data \cite{bishop2006pattern}. During training the generalization ability is approximated mainly by the empirical error. Particular design choices what the architecture of the DNN solution and the training procedure are concerned are made such as to further improve the generalization. These ensure good generalization despite infinite model capacity and limited training data. The validity of these design choices is typically investigated with the help of \textit{validation data} sampled from the unknown distribution independently of the development data set, which consists of the train and the test set. The DNNs are able to generalize as they learn a feature representation optimally suited to the classification problem they need to solve, assuming that the classification problem has a solution and the training data represents a proper sample.
\subsection{Concepts}
\label{Sec:literaturedir}

\begin{figure}[t]
	\centering
	\includegraphics[width=0.49\textwidth]{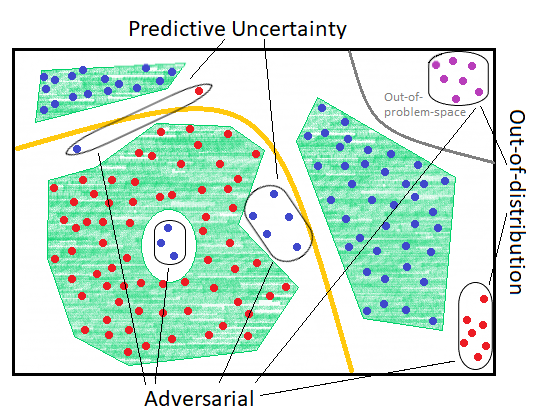}
	\caption{A feature space at inference time with a possible generalization area (green) for a classifier (orange) trained on a two class problem. For further details see section \ref{Sec:literaturedir}.}
	\label{Fig:Complement}
\end{figure}

Consider a classification problem in which, given separable and unambiguous realizations of several different concepts, we are supposed to group realizations of the same concept into a corresponding class. Assuming the realizations of concepts are observed by means of a system that produces images, which are available to us and we have to solve an image classification problem. For such a problem the corresponding \emph{problem space} consists of all images that depict valid realizations of concepts belonging to a predefined set-up such that each image can be meaningfully assigned to one of the classes. Such a set-up comes with an underlying distribution statistically describing the occurance of the images. 
\\
 The problem space usually consists of an infinite number of images. However, in order to develop a classifier a finite set of images is necessary. Therefore, a \emph{problem space sample} is drawn from the problem space. This problem space sample is used as the training data set for a classifier. We will focus on DNNs as classifiers. The classification DNN learns a \emph{feature space} and a classification boundary. In the feature space, the images can be separated more easily in their classes than in the vector space of the input image data. The aim is generalizing from the training set to any sample from the problem space distribution. The generalization error is estimated using a test set which usually is another subset sampled from the problem space, independent of the training set.
\\
\emph{The generalization envelope} of a DNN encloses the \emph{generalization area}, i.e., the region of the problem space for which the network decides reasonably correct on both previously seen and unseen input samples at inference time. A misclassified input sample at inference-time is called a generalization error.
\\
There are two main sources that can lead to a generalization error. One source is the model itself. In this case the architecture or the training procedure or both do not allow to learn a feature space and a decision boundary such that the data can be separated. A large enough DNN has infinite model capacity. Therefore the architecture and training procedure need to be chosen wisely such as to successfully generalize from the training sample and shatter the entire problem space. The second source of error stems from the training set, when this does not cover the problem space properly, in the sense that successful generalization is impossible. There may be several issues here: the problem-space sample may be sparse, it may cover only a region of the problem space, it may not reflect the distribution and/or extent of each concept correctly, etc.
\\
Consequently, the \emph{uncertainty} of a prediction can be separated into epistemic, which refers to uncertainty in the model predictions, and aleatoric, which captures uncertainty pertaining to the available problem-space sample. In practice it is often not possible to decide which kind of uncertainty caused a misclassification. The uncertainty information can not be entangled and therefore most uncertainty predictors are constructed to measure the combined uncertainty. Hence, in the following the term uncertainty includes epistemic as well as aleatoric uncertainty. As usual in the literature we we also use the term \textit{confidence} as an antonym for uncertainty, a high confidence implies a low uncertainty and vice versa. If for an input $x$, the DNN's output has a high uncertainty we assign a higher probability to the event that $x$ lies outside the generalization area. Conversely, a low uncertainty does not necessarily imply a smaller probability that the corresponding input lays outside the generalization area.  
\\
To illustrate our set-up, Figure \ref{Fig:Complement} depicts an example of a two dimensional feature space for a classification problem. The blue and red dots represent inference time data of two classes. A DNN is trained on a problem space sample and finds the orange decision boundary which splits the whole input space into two areas in which the data would be classified such that the area to the left corresponds to the red class and the data to the right corresponds to the blue class. The area in which the network is able to predict correctly is the generalization area visualized in green. Conversely, a correct decision given the combination of training data, classifier architecture and training process can not be guaranteed within the white area. As depicted in Figure \ref{Fig:Complement} there are typically three main types of data samples that are misclassified at inference time: Predictive uncertainty samples close to the separation surface, out-of-distribution samples located outside the problem space or far in the tail of the problem space distribution as approximated from the problem space sample and adversarial samples which are samples from within or outside the problem space purposefully selected or constructed such as a misclassification occurs, they therefore can be located in all regions outside the generalization area.
\\
 As visualized in Figure \ref{Fig:Complement} \textit{predictive uncertainty} samples are observed in regions close to a decision boundary. Such generalization errors are caused by a slightly misplaced decision boundary or noise in the data lead to a small misplacement of the sample in the feature space.  
 \\
\emph{Out-of-distribution} samples are often referred to as anomalies. They are different in a general way from the training samples. Either they are sampled from a part of the problem space not covered by the distribution estimated after training or they do not even belong to the problem space. Some possible out-of-distribution samples are marked at the right hand side of Figure \ref{Fig:Complement}. The purple color marks the fact that they are not in the problem space.
\\
The third class are \emph{adversarial examples}. They are constructed in order to fool the DNN by performing small changes on an image such that the sample is shifted to the wrong side of the decision boundary in the feature space which leads to a misclassification. Depending on their construction method adversarial examples can be found among out-of-distribution and predictive uncertainty samples, but also in \textit{low probability pockets} \cite{metzen2017detecting}, exemplarly visualized in the center of the red class area. Here, training data is sparse and therefore the region was not assigned correctly, i.e. to the blue class \footnote{Samples of incorrectly assigned low probability pockets do not have to be adversarial.}.\\
The literature on the detection of samples outside the generalization area may be divided into the described three main fields: \emph{predictive uncertainty, out-of-distribution detection and adversarial example detection}. Each field relates to a part of the question how to determine if an input has left the generalization area of a DNN. Further relations and specific technical definitions in each field are introduced and described respectively in Section \ref{Sec:probabilisitc}, \ref{Sec:OODMethods} and \ref{Sec:ADMethods}.

\subsection{Mathematical notation}
\label{Sec:Mathematical}
\begin{table}[t]
	\renewcommand{\arraystretch}{1.3}
	\caption{Nomenclature}
	\label{tab:nomenclature}
	\centering
	\begin{tabular}{|l | l| }
		\hline
		$x$~~~~~ & input data, in our case an image $x \in \mathbb{R}^{n}$\\
		\hline
		$l$~~~~~ & class label for an image $l \in \lbrace 1,\dots,m \rbrace$,\\
		& $m$ is the number of possible classes\\
		\hline
		$F(\cdot)$~~~~~ & DNN for classification $F:\mathbb{R}^n \rightarrow [0,1]^m$,\\
		& $x \mapsto F(x)~~\text{s.t.}~\sum_{i=1}^{m}F(x)_i=1$\\
		\hline
		$y$~~~~~ & predicted class $y \in \lbrace 1,\dots,m \rbrace$ \\
		& $y= \argmax_{i \in \lbrace 1,\cdots,m\rbrace} F(x)_i$\\
		\hline
		$L(\cdot,\cdot)$~~~ & loss function $L:[0,1]^m\times[0,1]^m\rightarrow \mathbb{R},$\\ 
		\hline
		&$(F(x),l) \mapsto L(F(x),l)$ compares the output\\ 
		&$F(x)$ of $x$ to the label $l$ to train the DNN \\
		\hline
		$\theta$~~~~~ & parameters of a DNN\\
		\hline
		$D_F(\cdot)$~~~~~~& detector $D_F(\cdot):\mathbb{R}^n \rightarrow \mathbb{R},~~ x \mapsto D_F(x)$ \\
		&predicts if $x$ is within ($D(F,x)<T$) \\
		&or outside ($D(F,x)\geq T $) the generalization\\
		& area of $F$; $T$ is a predefined threshold	\\
		\hline	
		
	\end{tabular}
\end{table}
For quick reference we have gathered in Table \ref{tab:nomenclature} the most important mathematical terms that we use. In the following we describe these in more detail.
\\
In the classification setup a DNN is a function $F(\cdot)$ that computes for an input $x\in  \mathbb{R}^n$ an output $F(x) \in \mathbb{R}^m$. The output $F(x)$ contains scores, one for each possible class. The maximal score defines the predicted class label $y$
$$y=\argmax_{i \in \lbrace 1, \dots,m \rbrace}{F(x)_i}\,.$$
Often, the last layer contains a softmax function that normalizes the final class scores
$$F(x) \in [0,1]^m,~~\text{s.t.}~\sum_{i=1}^{m}(F(x))_i=1\,.$$
Each DNN consists of several layers $f^j,~ j \in\lbrace1,...,k\rbrace$
$$F(x)=f^{(k)}(\dots f^{(2)}(f^{(1)}(x)))\,.$$
Usually each layer is a function on the output of the previous layer. Depending on the architecture and application area of the DNN, there are different kinds of layers, e.g. fully connected, convolutional or pooling layers. Often, the last layer contains a softmax activation. The parameters of these layers are the network's weights $\Theta$. The weights are optimized using a training set that contains input-output pairs $(x,l)$. The output labels $l\in \lbrace 1 ,\dots,m \rbrace$ hold the correct class for the inputs $x$. During training, the weights $\Theta$ are iteratively updated such that the loss $L(F(x),l)$, which compares the network output to the true label for samples from the training set, is reduced.
\\
The task for our work is to find a detector method $D_F(\cdot)$ that states at inference time for any possible input $x$ if it is within the generalization area of the DNN $F(\cdot)$. Based on the input $x$, the way the DNN processes $x$, and the output $F(x)$, the detector $D_F(\cdot)$ returns a value $D_F(x) \in \mathbb{R}$, which in combination with a threshold $T \in~ \mathbb{R}$ defines if $x$ is expected to be within the generalization area or not:
\begin{align*}
D_F(x)< T ~~\rightarrow &~x \text{ within generalization area}\\
D_F(x)\geq T ~~\rightarrow &~x \text{ not within generalization area}\,.
\end{align*}
Sometimes there is no hard decision needed, but the probability how likely the classification decision of the network is. This probability information can be achieved by defining a monotonously increasing function 
$$M: \mathbb{R} \rightarrow [0,1],$$ 
that maps the output of $D_F(x)$ onto the interval $[0,1]$. The closer $M(D_F(x))$ is to one the more likely the input is misclassified. The closer the value is to zero, the more likely the classification result of the DNN is correct.

\begin{figure*}[t]
	\centering
	\includegraphics[width=0.98\textwidth]{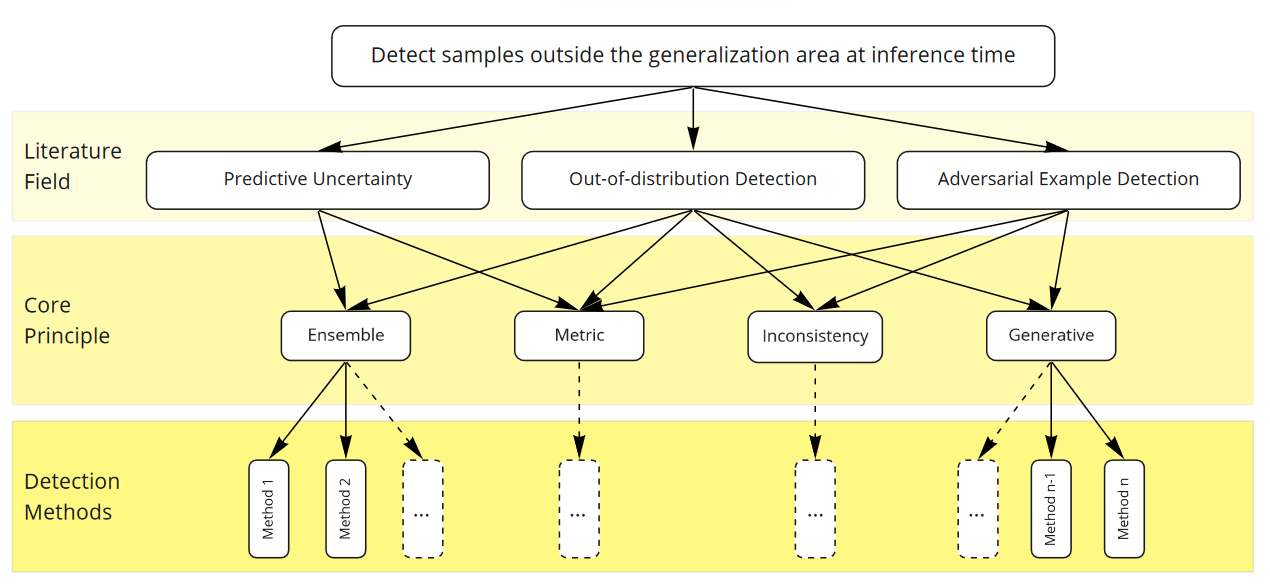}
	\caption{Taxonomy of methods detecting samples outside the generalization area at inference time.}
	\label{Fig:Structure}
\end{figure*}
\section{Related Work}
\label{Sec:Related}
To the best of our knowledge, there have been no other contributions focusing on the generalization capabilities of Machine Learning and in particular Deep Learning methods, as the root cause for research in the fields of, uncertainty estimation,  out-of-distribution detection and adversarial examples.
\\
Recently the most important methods in the area of uncertainty estimation have been benchmarked \cite{henne2020benchmarking}, \cite{ovadia2019can}.
\\ 
There are numerous surveys reviewing literature on out-of-distribution detection, which is also referred to as anomaly detection, especially in a context different from deep learning. Some review anomaly detection in general \cite{chandola2009anomaly}, \cite{pimentel2014review}. Others specialize on various setups such as data mining techniques \cite{agrawal2015survey} or on application areas as e.g. the medical domain or video-related anomaly detection\cite{kwon2017survey}, \cite{litjens2017survey}. In general those surveys that touch upon deep learning do not focus on anomalous behavior in combination with a predefined classifier \cite{chalapathy2019deep}, \cite{adewumi2017survey} and limit themselves to analyzing the training data. Detailed information on anomaly detection based on machine learning procedures is collected in several books \cite{dunning2014practical}, \cite{mehrotra2017anomaly}, \cite{aggarwal2015outlier}.
\\
Surveys on adversarial examples usually review literature on that topic in general and only include a section on adversarial example detection \cite{yuan2019adversarial}, \cite{akhtar2018threat}, \cite{chakraborty2018adversarial}.
\\ 
Bulusu et al. \cite{bulusu2020anomalous} focus on out-of-distribution and adversarial example detection methods applied on top of pre-trained DNNs. Different to our survey however, they do not use generalization to provide an unifying look on both fields and leave out predictive uncertainty.
\section{Detecting Samples Outside the Generalization Area}
\label{Sec:Methods}
In the following we introduce each literature field and discuss their detection methods thus setting the stage for a discussion focusing on the core principles they share (s. Section \ref{Sec:Combine}). As shown in Figure \ref{Fig:Structure} the core principles are: metric, inconsistency, generative and ensemble based methods. The various literature contributions within each category are arranged in ascending chronological order of their appearance.
\subsection{Predictive Uncertainty}
\label{Sec:probabilisitc}
For classification tasks the output values $F(x)$ of the DNN are usually understood as the corresponding probabilities of the sample $x$ belonging to class $y$. Against expectations it has been shown that this confidence score is misleading, as the DNNs tend to be overconfident \cite{goodfellow2014explaining}, \cite{nguyen2015deep}. Initial research on predictive uncertainty in the DNN-based classification context concentrated on adapting the output such that its values better approximate the actual probability distribution $p(y|x)$.
\\
The methods from the predictive uncertainty field are among the first attempts to investigate generalization at inference time, being older than out-of-distribution detection and adversarial example detection. Most procedures were originally constructed for regression problems, and were computationally cheap. Typically an ensemble of real world regression datasets first introduced by \cite{hernandez2015probabilistic} was used as evaluation set.
\\
Later, Ovadia et al. \cite{ovadia2019can} performed some image classification experiments on the most prevalent, scalable and practically applicable predictive-uncertainty deep-learning methods. To determine how well the predicted vector $F(x)$ fits the probability distribution of the corresponding probability $p(y|x)$ for each class $y$ different evaluation scores are used such as the \emph{Negative Log-Likelihood}, the \emph{Brier Score} and the \emph{Expected Calibration Error}.
\\
The \emph{Negative Log-Likelihood} for $N$ sample pairs $(x_i,l_i)$ is defined as
	$$NLL=-\sum_{i \in \lbrace 1,\dots,N \rbrace} log(F(x_i)_{l_i})\,.$$
It is also referred to as the cross entropy loss \cite{goodfellow2016deep}. It is a proper score, which means that the value $NLL$ is only minimized if the predicted probability distribution equals the groundtruth distribution.
 \\
Another proper score is the \emph{Brier Score} \cite{brier1950verification} which is for $N$ sample pairs $(x_i,l_i), ~ i \in \lbrace 1,\dots,N\rbrace$ given as
	$$BS=\frac{1}{N} \sum_{i \in \lbrace 1,\dots,N \rbrace} \sum_{j \in \lbrace 1,\dots,m \rbrace} (F(x_i)_j -\mathbf{1}_{\lbrace l_i \rbrace}(j))^2\,. $$
It can be explained as a decomposition between calibration and refinement, for further explanation we refer to \cite{degroot1983comparison}. 
\\
The \emph{Expected Calibration Error} \cite{naeini2015obtaining} is popular and intuitive. The evaluation data is split into $S$ disjoint buckets $B_s, ~s \in \lbrace 1, \dots,S \rbrace$ and the average gap of within bucket accuracy and within bucket predicted probability is measured 
	$$ ECE= \sum_{s \in {\lbrace 1,...,S \rbrace} }\frac{|B_s|}{n}|acc(B_s)-conf(B_s)|\,.$$
The accuracy and the confidence of a bucket $B_s$ with labeled samples $(x_i,l_i)$ and the corresponding predicted class $y_i, ~ i \in B_s$ is given as 
\begin{align*}
acc(B_s)=\frac{1}{|B_s| }\sum_{i\in B_s}\mathbf{1}_{\lbrace l_i \rbrace}(y_i)\\
conf(B_s)=\frac{1}{|B_s| }\sum_{i\in B_s} F(x)_{y_i}\,.
\end{align*}
The ECE is not a proper score, since e.g. returning the same value for all $F(x)_i, ~ i \in \lbrace 1,\dots,m\rbrace$ yields perfect calibrated, but non-usable predictions \cite{ovadia2019can}.\\
Next we explain the predictive uncertainty methods and group them according to their core principle into metric based and ensemble based approaches. \\
\subsubsection{Metric Methods} 

Metric-based methods investigate if for the current input data sample the classifier behaves similarly to the samples inside the generalization area. The training set is often used as the set of samples inside the generalization area. How similar the behaviors for two samples are is established by measuring the differences among them with the help of a function and comparing its output against a threshold. Typically the function used is a composition of some transformation and a metric. If a significant dissimilarity from the current sample to the samples from the inside of the generalization area is observed, the sample is expected to be outside the generalization envelope.\\

\noindent
The only metric method in the literature field predictive uncertainty was proposed by Guo et al. \cite{guo2017calibration}. Their simple method called temperature scaling "softens" the output $F(x)$ of the DNN. This is achieved by dividing the values of each class by a temperature $T>1$ before the softmax layer. The exact value of T is statistically derived from samples inside the generalization area. This procedure leaves the class prediction unchanged, since the parameter $T$ does not change the maximum, but the output values are softened, such that the output is less overconfident and hence more reliable. If the derived output value for the predicted class is much smaller than the output value for samples inside the generalization area, the sample is expected to be outside the generalization area.\\
\subsubsection{Ensemble Methods} 
\label{Sec:Ensemble} 
Ensemble methods usually apply the current input to several networks and analyse their outputs. The outputs are either combined to one output using an average procedure or the difference of the different outputs is computed. The more similar the components of the averaged output vector are, or the more the outputs of the individual networks differ, the more likely is the event that the investigated input sample is outside the generalization area. \\
\begin{figure}[t]
	\centering
	\includegraphics[width=0.45\textwidth]{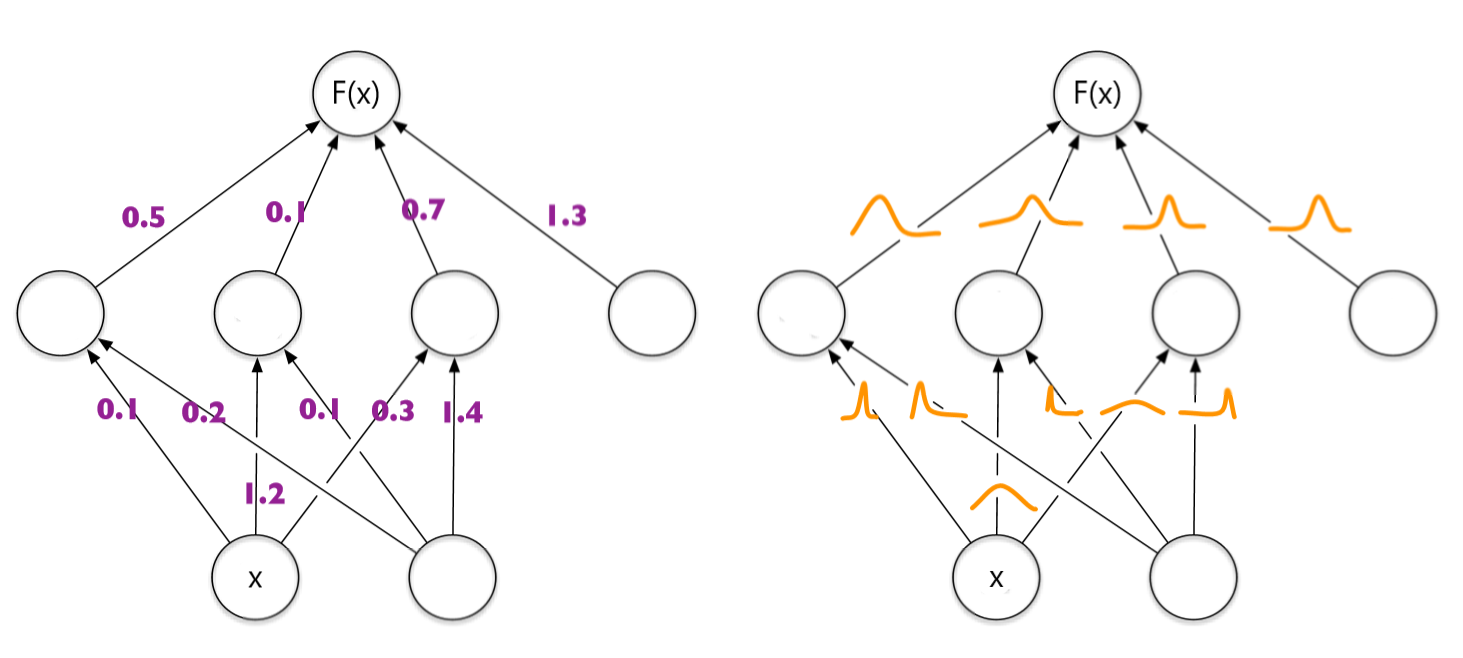}
	\caption{Visualization of the method from Blundell et al. \cite{blundell2015weight}.}
	\label{Fig:BNN}
\end{figure}

\noindent
One well known Bayesian Neural Network method is based on variational inference, in which an approximating distribution is constructed for each weight \cite{graves2011practical}, \cite{blundell2015weight}, \cite{louizos2016structured}, \cite{louizos2017multiplicative}, \cite{wen2018flipout}. Hence, the weights of the network are no longer fixed values but are represented by a probability distribution, compare Figure \ref{Fig:BNN}. One of the most popular Bayesian Neural Networks ideas is that from Blundell et al., since he was the first to introduce an algorithm for training Bayesian Neural Network using backpropagation \cite{blundell2015weight}. At inference time several concrete weight values are sampled from their corresponding distribution. The different samples are used to build an ensemble of networks. The input is run through each network and the outputs are combined using a weighted average. \\

\noindent
Gal et al. introduced a method that is called Monte Carlo dropout or dropout at inference time \cite{gal2016dropout}. Different from the standard dropout procedure \cite{srivastava2015training}, in which dropout is used to prevent overfitting by randomly deleting connections, in Monte Carlo Dropout this masking is also applied at test time. Hence, the prediction is no longer deterministic. Depending on the links randomly chosen to be kept, the networks output is different. For one input the network is run several times using different masks. The output results are averaged.\\

\noindent
Lakshminarayanan et al. also use an ensemble based method \cite{lakshminarayanan2017simple}. They randomly initialize several networks and train them on the same dataset. Additionally they use adversarial training, which means that they incorporate adversarial images in the training dataset to keep the models more robust against adversarial attacks. At inference time each DNN is run for the current input image and the output results are averaged. \\

\noindent
Riquelment et al. proposed two methods \cite{riquelme2018deep}. One relies on variational inference \cite{blundell2015weight}, the other on Monte-Carlo dropout \cite{gal2016dropout}. In order to enforce a lower computational overhead they incorporated the corresponding procedure only in the last layer of the network. Hence, the computation until the last layer stays the same for each input. Only in the last layer different weight settings need to be sampled and the corresponding output computed.\\
\subsection{Out-of-Distribution Detection}
\label{Sec:OODMethods}
Out-of-distribution detection methods search for input samples that stem from another distribution than the samples used to train the DNN. In the evaluation procedure additional datasets are used, which either contain inputs of different classes than the training classes or inputs with additional random noise. Some set-ups also include out-of-distribution images of classes the network is actually trained for but the images occur in a different representation. An example would be to feed images from the SVHN dataset \cite{netzer2011reading} which contains images of house numbers in a DNN that is trained to classify images from the handwritten digit dataset MNIST \cite{lecun1990handwritten}. Such cases are referred to as \emph{novelty detection} in literature and, often treated as subproblem of the out-of-distribution detection.
\\
Typically classification data sets used for out-of-distribution detection are CIFAR-10 and CIFAR-100 \cite{krizhevsky2009learning}, SVHN \cite{netzer2011reading}, ImageNet \cite{imagenet_cvpr09} and LSUN \cite{yu15lsun}. A DNN is trained on one of the datasets, the other datasets are then used as out-of-distribution data. 
\\
In the out-of-distribution detection set-up there are two classes, the inlier class covering the generalization area and the outlier class. A generalization detector returns for a sample $x$ and the corresponding behavior of the DNN a score. The higher the score the more likely the sample $x$ is an outlier, the lower, the more likely $x$ is an inlier. To get a hard decision for $x$ to be an inlier or an outlier, a decision threshold $T$ has to be determined.
\\
The ratio between False-Positive-Rate (FPR), and the True-Positive-Rate (TPR) is directly linked to the value $T$. However, depending on the problem and the corresponding requirements different ratios between the two rates can be desired and thus a threshold-independent evaluation metric is needed. For this purpose, the \emph{Area Under the Receiver Operating Characteristic (AUROC/AUCROC)} is typically used. This is the are under the plot of the true positive rate over the false positive rate. A 100\% AUROC value corresponds to a perfect detector, while a 50\%AUROC to a detector deciding at random. 
\\
Should the positive and negative classes have different base rates, the AUROC score can be misleading. In such a case it can make sense to use the \emph{Area under the precision recall curve (AUPR)}. It is defined as the area under the plot of the precision over the recall. Here, in contrary to the AUROC it makes a difference whether inliers or outliers are assigned to be negative or positive \cite{hendrycks2016baseline}.
\\
Lastly, sometimes the \emph{False positive rate at e.g. 95 \% true positive rate} is used. The threshold $T$ is chosen such that the true positive rate is at 95\% and the according value of the false positive rate is used for comparison.	
\\
Usually, the test data is adapted to have a ratio of one between the numbers of samples in the positive and negative class and the AUROC score is used for evaluation.
\\
The corresponding methods from the literature field out-of-distribution detection can be grouped into metric, inconsistency, generative and ensemble approaches. In the following, the methods are listed according to their group.
\subsubsection{Metric Methods}
\label{Sec:Metric}  
As already mentioned, metric methods determine if the behavior of the current input is similar to the behavior of samples inside the generalization area of the DNN under investigation. Again, the training samples are used as samples inside the generalization area and similarity is typically established using a transformation and a metric. In this case however, the transformation may be applied on the input data, the output of one or several intermediate layers or the gradient computed on the loss function of the current input-output combination. A large distance computed by the metric indicates a sample outside the generalization area.\\ 
\begin{figure}[t]
	\centering
	\includegraphics[width=0.48\textwidth]{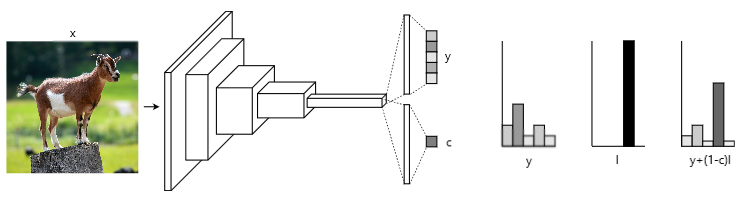}
	\caption{Visualization of the method from DeVries et al., image adapted from \cite{devries2018learning}.}
	\label{Fig:LearningConfidenceModel}
\end{figure}

\noindent
DeVries et al. proposed a method in which an usual classification DNN is adopted such that it outputs a detection value $c$ additionally to the softmax scores \cite{devries2018learning}, compare Figure \ref{Fig:LearningConfidenceModel}. To train the network they use a two-folded loss function $\tilde{L}$. On the one hand it interpolates the softmax output $y$ with the true class one-hot label $l$ and hence lowers the classification loss if $c$ is low, and on the other hand it is penalizing a small confidence score $c$
$$\tilde{L}((y,c),l)= L(y+(1-c)\cdot l,l)+ log(c)\,.$$
Via this training process the detecting transformation and the metric based on the layerwise output of the DNN is directly incorporated in the network structure which is different to most other metric methods. At inference time a small value for $c$ indicates that the sample is likely to be outside the generalization area.\\

\noindent
Oberdiek et al. introduced a method that is based on the layer-wise gradient of the weights regarding the loss function of the predicted class \cite{oberdiek2018classification}. From that they generate features such as the layerwise norm, minimum and maximum values of the gradient. Those are fed together with the entropy of the estimated class distribution to a logistic regression approach that determines the score deciding if the input is expected to be outside the generalization area.\\

\noindent
Jiang et al. proposed a method that first defines for each class a high density set consisting of images from the training set of that according class \cite{jiang2018trust}. At inference time a \emph{trust score} for the current input output combination is computed by taking the ratio between the distance from the test sample to the high density class set of the predicted class and the distance to the second nearest high density class set. The higher the trust score, the more the predicted class is expected to be correct.\\

\noindent
Another detector that is completely based on the activation spaces of the DNN's layers was introduced by Lee et al. \cite{lee2018simple}. They compute the Mahalanobis distance to the closest class-conditional Gaussian distribution. The layerwise distances are then combined by a logistic regression network stating if the input is likely to be outside the generalization area. It is one of the only works yet that evaluates the detector on both, adversarial examples and out-of-distribution set-ups.\\

\noindent
Hendrycks et al. adapts the training procedure of the DNN \cite{hendrycks2019deep}. They carefully construct a dataset contains samples outside the generalization area that is different from that used for testing. During training they use the original training set with the original labels and additionally samples from the constructed dataset for which they use the uniform distribution over $m$ classes as labels. An input leading to an output with an higher entropy than samples from the training set within the generalization area are expected to be outside the generalization area.\\

\noindent
A similar method was proposed by Hein et. al \cite{Hein_2019_CVPR}. They also use an additionally constructed dataset with samples outside the generalization are for the training. During training the loss for such samples is defined by the maximal output value over all classes. During testing a comparison of the maximal output value to that of common samples inside the generalization area is used to detect samples outside the generalization area.
\subsubsection{ Inconsistency Methods} 
\label{Sec:Inconsistency} 
The core idea of prediction-inconsistency methods is to observe the reaction of the DNN to small changes in the input image, when the input image lays inside and respectively outside the generalization area. At inference time the current input sample and a slightly transformed image are processed by the DNN. The corresponding outputs are compared. The higher the difference in the outputs the more the investigated sample is expected to be outside the generalization area.\\

\noindent
Liang et al. proposed a method in which the input image is perturbed by shifting it away from the original class \cite{liang2017enhancing}. This is done similarly as in the adversarial example method FGSM \cite{goodfellow2014explaining}: gradient descent is used to increase the loss regarding the predicted class and hence a slightly modified image is generated. A sample is expected to be inside the generalization area, if the output value of the modified image for the original predicted class is high. 
\subsubsection{Generative Methods} 
\label{Sec:Generative} 
Generative methods make use of an image pre-processing procedure. The idea is somewhat similar to the prediction-inconsistency approaches, but the pre-processing procedure is more sophisticated since it is generated in order to shift the image in the direction of the training distribution. The shift is realized using a generative network or similar which is trained on the training set. At inference time both the current input image and the corresponding output of the generative network (i.e., the result of applying the pre-processing procedure on the input image) are processed by the DNN. The more the outputs differ, the more the input sample is assumed to be outside the generalization area.\\
\begin{figure}[t]
	\centering
	\includegraphics[width=0.3\textwidth]{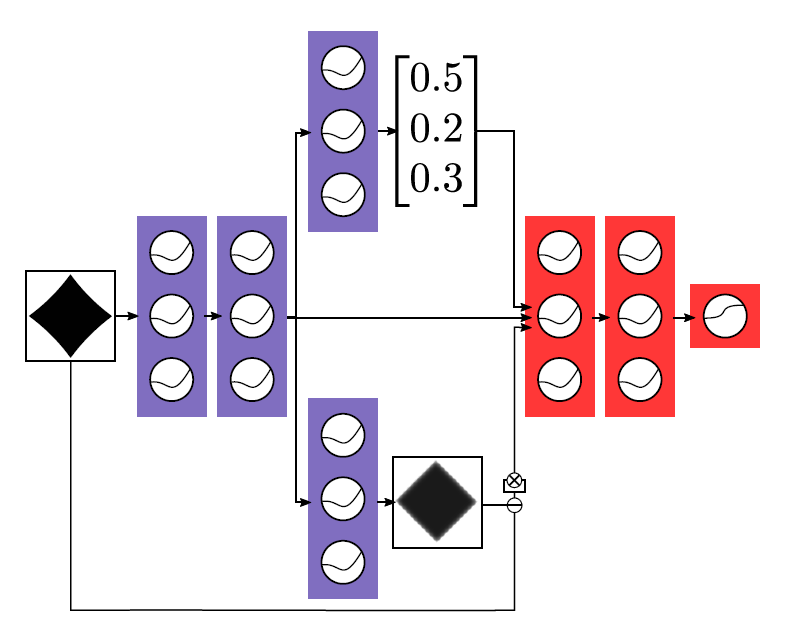}
	\caption{Visualization of the Method from Hendrycks et al. \cite{hendrycks2016baseline}.}
	\label{Fig:baselineOut}
\end{figure}

\noindent
Hendrycks et al. proposed a method that requires changes in the original DNN \cite{hendrycks2016baseline}. As shown in Figure \ref{Fig:baselineOut} in purple, they attach an encoder on top of the penultimate layer of the original DNN, that reconstructs the input image. The difference between the encoder output and the input image is then fed together with the output of the penultimate and last layer to a so called abnormality module shown in red which is trained to output the generalization score for the prediction.\\

\noindent
Ren et al. introduced a method that is based on a likelihood ratio statistic \cite{ren2019likelihood}. They train two generative models called PixelCNNs \cite{salimans2017pixelcnn++} on training samples that additionally returns how likely an input is side the generalization area. One model is trained on the original training dataset and the other one on slightly perturbed images from the training dataset. They suppose that both models are able to capture background but that the model trained on the original training dataset is more sensitive regarding the actual content part. Hence, if the likelihood ratio determined between the output of the model trained on the original images and the one on the perturbed images is low, the input image is expected to be outside the generalization area.\\

\noindent
Serr\`{a} et al. show that images with high complexity tend to produce lowest likelihoods for generative models \cite{Serra2020Input}. This leads to wrong predictions if the sample outside the generalization area has a lower complexity than the actual data inside the generalization area. They balance this effect in a method, that takes both the predicted log-likelihood of generative models and the quantitative estimates of complexity of an image into account to decide whether this image is outside the generalization area. They use several different generative models in order to evaluate their method.
\subsubsection{Ensemble Methods} As described in Section \ref{Sec:Ensemble}, ensemble methods use several slightly different networks, the more the outputs of those networks differ at inference time, the more the image is expected to be outside the generalization area. \\

\noindent
Vyas et al. splits the training data in $k$ partitions, such that each class is belonging to one partition \cite{vyas2018out}. Then $k$ independent but structure wise identical networks are trained, each uses one of the partitions as data outside the generalization area and the rest as in-distribution data. Additional to the usual loss term, they use a term that pushes the entropy over the softmax output of the network below a threshold for samples outside the generalization area and above the threshold for in-distribution samples. During testing the softmax outputs of the classifiers are averaged and the maximum value and the entropy are combined and used as detection score. \\

\noindent
Another ensemble based method was proposed by Yu et al., they are training two networks, both on the same in- and out-of-distribution examples \cite{yu2019unsupervised}. For in-distribution images they use an usual loss function, but for out-of-distribution examples they use a loss that is thought to maximize the difference between the softmax outputs of the two networks. During testing they expect the input to be outside the generalization area if the L1-Norm of the difference of the outputs of the two networks is above a predefined threshold. 

\subsection{Adversarial Example Detection}
\label{Sec:ADMethods}
\begin{figure}[t]
	\centering
	\includegraphics[width=0.48\textwidth]{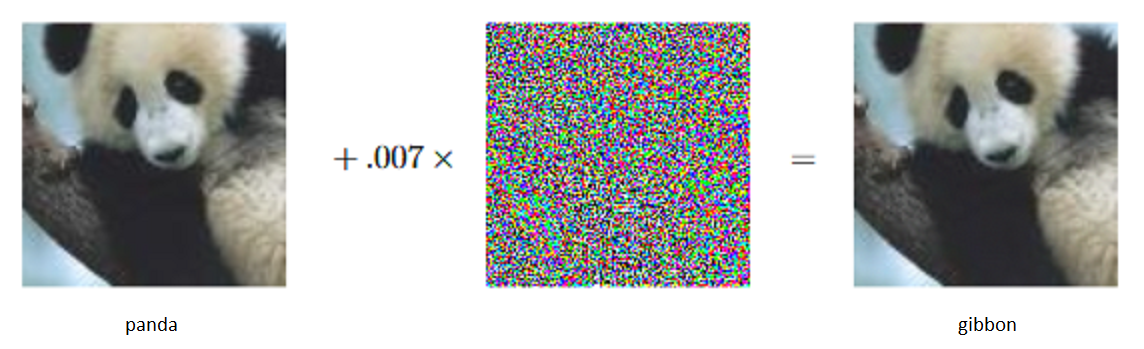}
	\vspace{+0.2cm}
	\caption{An original "clean" image that is correctly predicted as panda, the added perturbation computed by the fast gradient sign method and the resulting adversarial example which is incorrectly classified as gibbon, image adapted from \cite{goodfellow2014explaining}.}
	\label{Fig:fgsm}
\end{figure}

Adversarial examples are artificially generated data samples that are carefully constructed to fool a network into deciding falsely. Usually, an adversarial example $x_{adv}$ is based on an original data sample $x$. The difference $\delta$ between those samples is constructed to be as small as possible under the constraint that the predicted class $c(F(x))$ changes
$$x_{adv}= \argmin_{x_{adv}} ||x-x_{adv}||~~~~ s.t.~~~~ c(F(x)) \neq c(F(x_{adv}))\,. $$
The difference between the two samples is often not visible to humans, as shown in Figure \ref{Fig:fgsm}. 
This adversarial example, for which the network predicts a gibbon instead of a panda is constructed using the \emph{fast gradient sign method (FGSM)} by Goodfellow et al. \cite{goodfellow2014explaining}, which is one of the first and simplest adversarial attack methods. Instead of computing the gradient of the loss function $L(F(x),l)$ regarding the weights $\theta$ of a model, the gradient regarding the image $x$ is computed. Then the adversarial image $x_{adv}$ is generated by moving the original image $x$ in a simple one step procedure in the direction of the steepest ascent regarding the loss $L(F(x),l)$, without exceeding the L1 norm difference $\epsilon$ to the original image. 
$$x_{adv}=x + \epsilon \cdot sign( \nabla_x L(\theta,x,y))\,.$$   
\\
The main adversarial example methods used for evaluating adversarial example detectors are FGSM \cite{goodfellow2014explaining}, BIM \cite{kurakin2016adversarial}, JSMA \cite{papernot2016limitations} and CW \cite{carlini2017towards}. Typical datasets used for evaluation are MNIST \cite{lecun1990handwritten}, CIFAR \cite{krizhevsky2009learning} and SVHN \cite{netzer2011reading}. In order to prove the scalability to large datasets some works additional evaluate their methods on the ImageNet dataset \cite{imagenet_cvpr09}. In the adversarial sample set-up usually the DNN is trained on the training data. At inference time the DNN receives both, adversarial and clean images from the test set. The task of the generalization detector is to decide whether the pre-trained DNN  has as input an adversarial example from outside the generalization envelope and hence predicts the wrong class, or if the current input data is clean and the network predicts correctly.\\
The evaluation procedure for adversarial example detection is similar to the evaluation of out-of-distribution detection explained in Section \ref{Sec:OODMethods}, with the adversarial examples being considered outliers. 
\\
The methods for adversarial example detection can be grouped into metric, inconsistency and generative. In the following the methods are explained and listed according to their core principle. 
\subsubsection{Metric Methods} As described in more detail in Section \ref{Sec:Metric} the metric methods use the information if the current input sample is behaving similar to input samples inside the generalization area that are taken from the training set. It is based on a transformation and a metric applied on the input data, on the output of one or several layers or on the gradient computed on the loss function of the current input output combination. A high difference in the behavior indicates an adversarial sample laying outside the generalization area.\\

\noindent
Grosse et al. introduced an approach that applied maximum mean discrepancy on input data \cite{grosse2017statistical} from which a distance to samples inside the generalization area is computed. This was one of the first works for the detection of adversarial examples. The main difference to the following works in this category is that the procedure only depends on the images, the information is independent from the behavior of the DNN. It later showed that this method does not work well for some adversarial attacks, hence following methods in the metric based category took more information than just the image space itself into account.\\

\begin{figure}[t]
	\centering
	\vspace{+0.3cm}
	\includegraphics[width=0.5\textwidth]{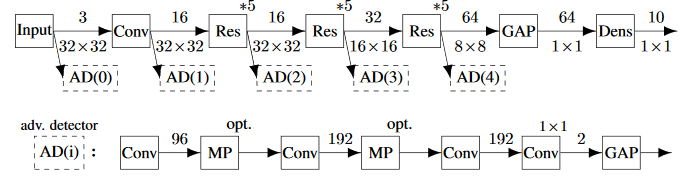}
	\vspace{+0.1cm}
	\caption{Subnetworks for the detection task trained on the output of the activation space. Original image from Metzen et al. \cite{metzen2017detecting}.}
	\label{Fig:metzenDetecting2017}
\end{figure}

\noindent
Metzen et al. proposed one of the first detectors that uses the activation space of the DNN as decisive criterion \cite{metzen2017detecting}. On top of each activation layer output they trained small subnetworks that receive the activations of the original network for each example as input. The subnetworks were trained on data in- and outside the generalization area. A visualization of this method can be seen in Figure \ref{Fig:metzenDetecting2017}. Based on that information the subnetworks classify if the current example is outside the generalization area. Metzen et al. found that the subnetwork on one of the middle layers leads to the best detection results.\\

\noindent
Li and Li developed a detector that is using the output of the convolutional layers in the DNN \cite{li2017adversarial}. For each filter output it collects statistical information from normalized principal component analysis, minimal and maximal values and several percentile values. Based on those statistics a distance is computed to samples inside the generalization area from the training set and a decision is made if the current input behaves similar to previously investigated image samples.\\

\noindent
A two-folded detector was introduced by Feinman et al. \cite{feinman2017detecting}. The main procedure is based on the Kernel Density calculated on the activation space of each layer which is then compared to samples of the training set. The second procedure uses Bayesian Uncertainty determined from several runs using dropout in the DNN. If one of the detectors determines deviations from a normal behavior, the current example is detected as being outside the generalization area.\\

\noindent
A procedure based on local intrinsic dimensionality and was developed by Ma et al. \cite{ma2018characterizing}. They characterize subspaces outside the generalization area using local intrinsic dimensionality on the layer outputs of the DNN which is a weighted distance metric computed on the k-nearest-neighbors of 100 randomly chosen examples from the training set. The results of the different layers are then combined by a logistic regression network.\\

\noindent
A method also based on a k-nearest-neighbor approach was proposed by Papernot et al. \cite{papernot2018deep}. In each layer's activation space a input image of the k-nearest neighbor images of the training set are determined. This method was actually thought and is inserted as an example for several methods that actual recover the actual class of the adversarial image but could be easily updated: if the classes mismatch to often with the predicted class of the current input image, the image is likely to be outside the generalization area.\\

\noindent
Zheng et al. proposed a method that fits for each class on each activation space of the fully connected layers a Gaussian Mixture Model \cite{zheng2018robust}. The Gaussian Mixture Model is trained by an expectation-maximization algorithm \cite{mclachlan2007algorithm}. For each class a threshold is chosen to reject inputs whose Gaussian Mixture Models values are below that threshold.\\

\noindent
Pang et al. introduced a method based on the entropy of normalized non-maximal elements (non-ME) of the predictions of the network outputs $F(x)$ \cite{pang2018towards}
$$\text{non-ME}(x)=-\sum_{i\neq y}F(x)_i\cdot log(F(x)_i)\,.$$
The training of the DNN is adapted with a loss function that has an additional non-ME term, to keep the non-ME value for samples outside the generalization area low. During testing, an image $x$ with a high non-ME value is expected to be outside the generalization area.\\

\noindent
As mentioned in section \ref{Sec:OODMethods}, the method proposed by Lee et al. evaluated on the detection of adversarial examples as well as out-of-distribution data \cite{lee2018simple}. It is one of the only works yet, that evaluates the detector on both set-ups.\\
\begin{figure}[t]
	\centering
	\includegraphics[width=0.48\textwidth]{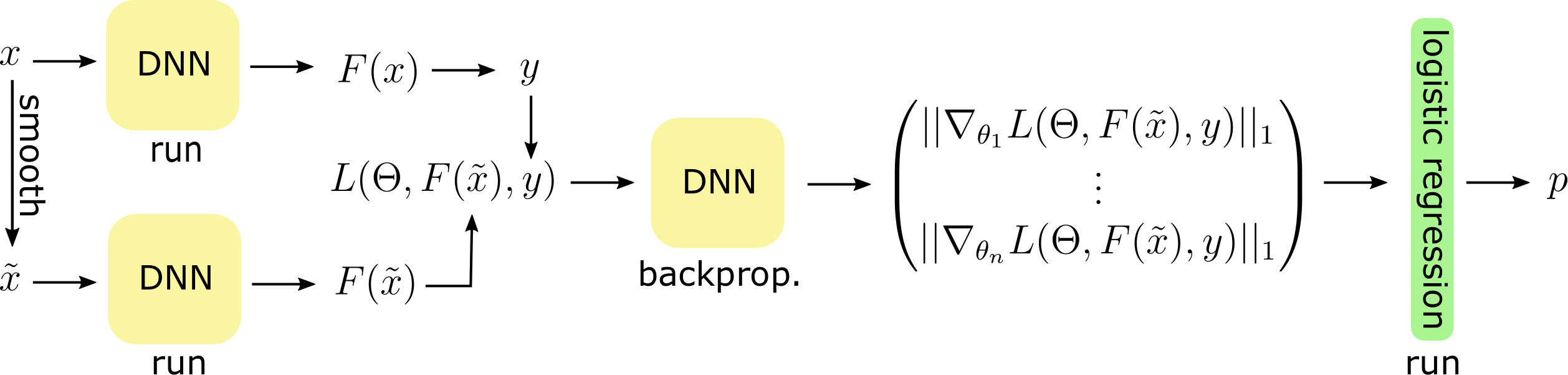}
	\caption{An illustration of the gradient based detector method of Lust and Condurache \cite{lust2020gran}.}
	\label{Fig:Gran}
\end{figure}

\noindent
Ma et al. introduced an approach that is on the one hand based on a one-class support vector machine on the activation layers, and on the other hand they apply on top of each layer an additional fully connected softmax layer which directly returns a softmax score class vector \cite{ma2019nic}. These vectors are compared to each other. The more they differ, the more probable the current example is misclassified. Furthermore, the output of the support vector machine is taken into account. The combined information decides whether the current input is expected to be outside the generalization area.\\

\noindent
A similar procedure to Papernot et al.'s method \cite{papernot2018deep} was proposed by Dubey et al. \cite{dubey2019defense}. It is using only a few feature representations constructed from the activations from different intermediate layers of the input image. Those feature representations are compared to pre-saved representations of labeled images from a large database. Nearest neighbor search is used, followed by a weighted combination of the predictions of the nearest neighbors leading to a final class prediction. The approach is original thought to recover the original class, but it can be adapted to a detector and the image is supposed to be outside the generalization area if the class predictions mismatch.\\

\noindent
Recently Lust and Condurache introduced a detector based on the gradient of the weights regarding the loss function that compares the predicted class to the softmax output of the network \cite{lust2020gran}. They use the layerwise norm of the gradient as features in the logistic regression detector. Furthermore, they add a smoothing step in order to remove perturbing noise and hence, in case of a sample outside the generalization area, increase the contradictions of the weights to the predicted class, compare Figure \ref{Fig:Gran}.
\subsubsection{Inconsistency Methods} As described in Section \ref{Sec:Inconsistency} prediction-inconsistency methods use the difference in the DNN output when processing an input sample and a slightly modified input, between samples that lay inside and outside the generalization area. A large difference indicate a sample outside the generalization area. \\

\noindent
One of the first methods in this category was proposed by Liang et al. \cite{liang2018detecting}. They aim to remove adaptive noise from the current input image and hence the perturbation by the use of scalar quantization and a spatial smoothing filter. The current input image and the generated clear image are then classified by the DNN. If the class prediction for the two is different, the image is supposed to be outside the generalization area.\\
\begin{figure}[t]
	\centering
	\includegraphics[width=0.5\textwidth]{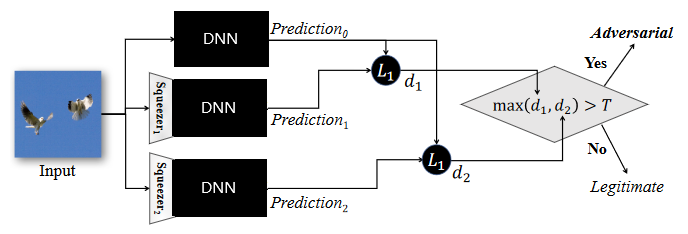}
	\caption{Visualization of the prediction inconsistency based approach of Xu et al. \cite{xu2018feature}.}
	\label{Fig:featuresqueezing}
\end{figure}
\noindent
The most successful approach in the prediction inconsistency based approaches is if from Xu et al. and called \emph{feature squeezing} \cite{xu2018feature}. The original image, an additional image with reduced color depth generated from the original image and a smoothed image are run through the DNN. The outputs of the generated images are compared to the output of the original image. If one of those differences exceeds a threshold, the image is detected as an image outside the generalization area.
\subsubsection{Generative Methods}
As described in Section \ref{Sec:Generative} generative based methods shift the input image in the direction of the training distribution. A high difference in the outputs of the current input image and the corresponding shifted image indicates that the input image is outside the generalization area.\\

\noindent
Meng et al. introduced MagNet, a two-pronged detection method \cite{meng2017magnet}. They use an auto-encoder that is trained to reconstruct examples of the original dataset. Each input of the DNN is additionally processed by the autoencoder and a reconstructed image is achieved and as well run through the DNN. If the current sample and its reconstruction differ a lot or the difference between the DNN's output of the current image and the DNN's output of the reconstructed image is too high, the sample is supposed to be outside the generalization area.\\

\noindent
Song et al. proposed a method that is based on PixelCNN \cite{ord2016pixel}, \cite{salimans2017pixelcnn++}, a generative model with tractable likelihood \cite{song2017pixeldefend}. They train this PixelCNN on the original training data. During testing each image is run through a procedure in which each pixel of the image is modified such that the log-likelihood of the PixelCNN is maximized. This procedure is thought to remove the perturbation. If the class prediction of the "cleaned" image is different to the class prediction of the original image, the image is supposed to be outside the generalization area. \\

\noindent
\emph{Defense-gan} is a method proposed by Samangouei et al. \cite{samangouei2018defense}. They train a Generative adversarial network (GAN) \cite{goodfellow2014generative} on the original dataset. Next they find an input $z$ to the generator part $G$ of the GAN via an iterative procedure, such that the output image $G(z)$ is as similar as possible to the current input image $x$ of the DNN. The image $G(z)$ is then also fed into the DNN. Again, a difference in the predicted class leads the method to predict the input to be outside the generalization area.\\
\begin{figure}[t]
	\centering
	\includegraphics[width=0.3\textwidth]{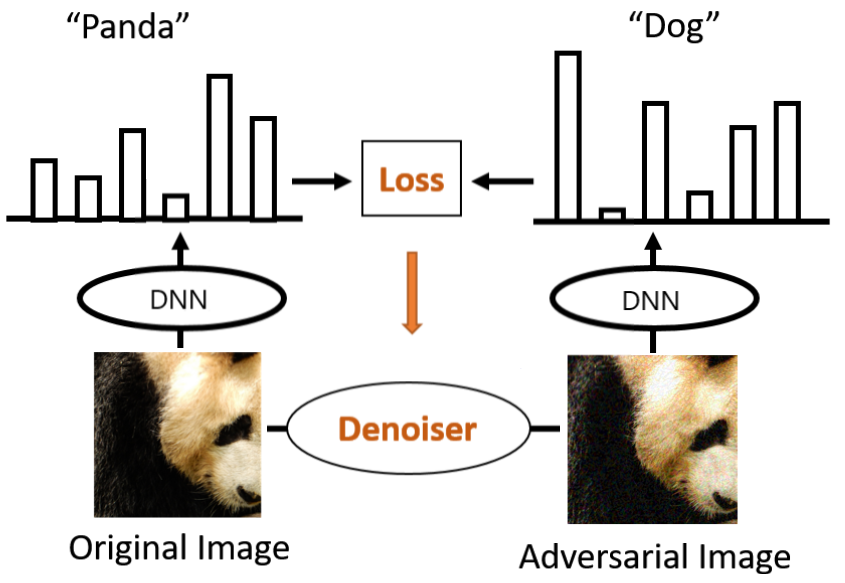}
	\caption{The detector method of Liao et al. using a high-level representation guided denoiser \cite{liao2018defense}.}
	\label{Fig:HGD}
\end{figure}

\noindent
Liao et al. proposed a similar detector method to MagNet. They use a high-level representation guided denoiser \cite{liao2018defense} which is instead of the typical encoder-decoder structure based on a UNet structure to generate the reconstructed image. It is trained on clean and adversarial images such that adversarial images have the same DNN top-level outputs as the corresponding non adversarial samples. The image is classified as being outside the generalization area if the class output of the denoised and the original image differs. The method is visualized in Figure \ref{Fig:HGD}. 
\section{Discussion}
\label{Sec:Discussion}
In the following we discuss and summarize advantages and disadvantages of the introduced methods. Important comparison criteria are listed in Table \ref{tab:comparison}. In the first section, we clarify the used criteria and discuss the findings in the second section.
\begin{table*}[t]
	\renewcommand{\arraystretch}{1.33}
	\caption{Compact comparison of detection methods. For further details on the comparison criteria see section \ref{Sec:Criteria}.}
	\label{tab:comparison}
	\centering
	\small
	{
		\begin{tabular}{|l|c|c|c|c|c|}
			\hline
			\multirow{2}{*}{Method reference}	 & Core    		& No link to  & No outlier     & Additional  \cmmnt{& Code} &  Publication    \\
			&  principle    & inference engine & modeling &  parameters \cmmnt{& available}  &  date \\
			\hline
			\hline
			\multicolumn{6}{|l|}{Literature Field:  Predictive uncertainty methods} \\
			\hline
			
			Guo et al. \cite{guo2017calibration}&Metric& \cmark & - & \Circle \cmmnt{& \cmark} &08.2017 \\
			Blundell et al. \cite{blundell2015weight} & Ensemble & \xmark & -& $~\mathrlap{\Circle}\LEFTCIRCLE$ \cmmnt{& \cmark} & 06.2015 \\
			Gal et al. \cite{gal2016dropout} & Ensemble& \xmark&- & \Circle\cmmnt{& \cmark} &06.2016 \\
			
			Lakshminarayanan et al. \cite{lakshminarayanan2017simple}&Ensemble&\xmark & - & $\CIRCLE$ \cmmnt{ &\cmark }&12.2017\\
			Riquelme et al. \cite{riquelme2018deep}&Ensemble& \xmark & -&\Circle \cmmnt{& \cmark }&05.2018\\
			
			\hline
			\hline
			\multicolumn{6}{|l|}{Literature Field:  Out-of-distribution detection} \\
			\hline
			DeVries et al. \cite{devries2018learning}&Metric&\xmark&(\cmark)&\Circle\cmmnt{&\cmark }&02.2018\\
			Oberdiek et al. \cite{oberdiek2018classification}&Metric&\cmark&(\cmark)&\Circle\cmmnt{&-}&09.2018\\
			Jiang et al. \cite{jiang2018trust}&Metric&\cmark&\xmark&\CIRCLE\cmmnt{&\cmark} &12.2018\\
			Lee et al. \cite{lee2018simple}&Metric&\cmark&(\cmark)&$\CIRCLE$\cmmnt{&\cmark }&12.2018\\
			Hendrycks et al. \cite{hendrycks2019deep} & Metric  & \xmark & (\cmark) & \Circle \cmmnt{&\cmark}& 04.2019 \\
			Hein et al. \cite{Hein_2019_CVPR}& Metric & \xmark & (\cmark) & \Circle \cmmnt{&\cmark}& 06.2019 \\
			Liang et al. \cite{liang2017enhancing}&Inconsistency&\cmark&\xmark&\Circle\cmmnt{&\cmark} &04.2018\\
			Hendrycks et al. \cite{hendrycks2016baseline}&Generative&\xmark&(\cmark)& $\mathrlap{\Circle}\LEFTCIRCLE$\cmmnt{&\cmark} &10.2016\\
			Ren et al. \cite{ren2019likelihood}&Generative&\xmark&(\cmark)& $\mathrlap{\Circle}\LEFTCIRCLE$\cmmnt{&-}&12.2019\\
			Serr\`{a} et al. \cite{Serra2020Input}&Generative&\cmark & \cmark & $\mathrlap{\Circle}\LEFTCIRCLE$\cmmnt{&- }&04.2020\\
			Vyas et al. \cite{vyas2018out}&Ensemble&\xmark&\cmark&\CIRCLE\cmmnt{&-}&09.2018\\
			Yu et al. \cite{yu2019unsupervised}&Ensemble&\xmark&\xmark&$\mathrlap{\Circle}\LEFTCIRCLE$\cmmnt{&-}&10.2019\\
			\hline
			
			
			\hline
			\hline
			\multicolumn{6}{|l|}{Literature Field:  Adversarial example detection} \\
			\hline
			Grosse et al. \cite{grosse2017statistical}&Metric&\cmark&\cmark&$\mathrlap{\Circle}\LEFTCIRCLE$\cmmnt{&-}&02.2017\\
			Metzen et al. \cite{metzen2017detecting}&Metric&\cmark&(\xmark)&\CIRCLE\cmmnt{&-}&04.2017\\
			Li and Li \cite{li2017adversarial}&Metric&\cmark&\xmark&$\mathrlap{\Circle}\LEFTCIRCLE$\cmmnt{&-}& 10.2017\\
			Feinman et al. \cite{feinman2017detecting}&Metric&\cmark&\xmark&\CIRCLE\cmmnt{&-}&11.2017\\
			Ma et al. \cite{ma2018characterizing}&Metric&\cmark&\xmark&\CIRCLE\cmmnt{&\cmark}& 03.2018\\
			Papernot et al. \cite{papernot2018deep}&Metric&\cmark&\xmark&\CIRCLE\cmmnt{&-}&03.2018\\
			Zheng et al. \cite{zheng2018robust}&Metric&\cmark&\cmark&$\mathrlap{\Circle}\LEFTCIRCLE$\cmmnt{&-}&05.2018\\
			Pang et al. \cite{pang2018towards}&Metric&\xmark&\xmark&$\mathrlap{\Circle}\LEFTCIRCLE$\cmmnt{&\cmark }&12.2018\\
			Lee et al. \cite{lee2018simple}&Metric&\cmark&(\cmark)&$\CIRCLE$\cmmnt{&\cmark}&12.2018 \\
			Ma et al. \cite{ma2019nic}&Metric&\cmark&\cmark&\CIRCLE\cmmnt{&-}&02.2019\\
			Dubey et al. \cite{dubey2019defense}&Metric&\cmark&\cmark&\CIRCLE\cmmnt{&-}&03.2019\\
			Lust and Condurache \cite{lust2020gran}&Metric&\cmark&\xmark&\Circle\cmmnt{&-}&04.2020\\
			Liang et al. \cite{liang2018detecting}&Inconsistency&\cmark&(\xmark)&\Circle\cmmnt{&-}&04.2018\\
			Xu et al. \cite{xu2018feature}&Inconsistency&\cmark&(\xmark)&\Circle\cmmnt{&\cmark}&02.2018\\
			Meng et al. \cite{meng2017magnet}&Generative&\cmark &\cmark &$\mathrlap{\Circle}\LEFTCIRCLE$\cmmnt{&-}&05.2017\\
			Song et al. \cite{song2017pixeldefend}&Generative&\cmark&\cmark&$\mathrlap{\Circle}\LEFTCIRCLE$\cmmnt{&\cmark}&10.2017\\
			Samangouei et al. \cite{samangouei2018defense}&Generative&\cmark& \xmark & $~\mathrlap{\Circle}\LEFTCIRCLE$ \cmmnt{&\cmark} &05.2018\\
			Liao et al. \cite{liao2018defense}&Generative&\cmark&\xmark&$\mathrlap{\Circle}\LEFTCIRCLE$\cmmnt{&\cmark} &06.2018\\
			\hline
			
		\end{tabular}
	}
\end{table*}
\subsection{Comparison Criteria}
\label{Sec:Criteria}
Table \ref{tab:comparison} captures the methods described in Section \ref{Sec:Methods} ordered by their literature field and core principle.
 \\
The third column, \textbf{no link to inference engine}, defines whether a method can be directly applied to any DNN. If yes, the method belongs to the class of post-hoc procedures, which means that a pre-trained DNN can be taken and no further adaption is needed, this is marked by \cmark . If e.g. a special loss function or additional layers needs to be incorporated in the classification DNN and hence the training process needs to be adapted, the method is marked with an \xmark.
\\
The column labeled \textbf{no outlier modeling} states if training samples from the outlier class, e.g. adversarial or out-of-distribution samples are necessary to train the detector method. Here, we distinguish between 4 categories. \cmark~means the training of the method is completely outlier independent. If the method is marked by (\cmark) outliers are necessary for the training of the method but the method is (additionally) evaluated on a set-up in which the training outliers were sampled from a different distribution than the test outliers, the method was able to generalize well. (\xmark) means that they evaluate (additionally) on a set-up in which the training outliers were sampled from a different distribution than the test outliers but the method was not able to generalize well. Methods that are evaluated only on outlier set-ups they were in particularly trained for are marked by \xmark.
\\
In the \textbf{additional parameter} comparison we group the methods into three subgroups. The first group consists of detector methods that do not need many additional parameters \Circle ~in comparison to the classification DNN. If the number of additional parameters is in the range of the numbers of parameters needed for the DNN the methods belong to the second group $\mathrlap{\Circle}\LEFTCIRCLE$. Lastly, if the number of parameters needed for the detection method is more than twice the numbers of parameters needed for the DNN they belong to the third group \CIRCLE.
 \\
 For most methods the \textbf{computational overhead} is not given within the corresponding paper and we did not do any experiments on that. Due to the lack of precise information we were not able to include an extra column for the computational efficiency in the table. However, some rough estimations can be made. The most time consuming methods are most likely the ensemble based approaches, since here all the slightly different networks, often two or more, have to process the same input. For the other categories we expect the computational overhead to be in the range of the parameter overhead, since usually one parameter is related to one computation step. Thus, the only methods needing significant fewer parameters and computational overhead than the original network itself are found within the metric based approaches category. This information is relevant for applications in hardware restricted areas as e.g. autonomous driving \cite{gauerhof2018structuring}, \cite{feng2018towards}.
 \\
The last column named \textbf{publication date} gives the month and the year the method was introduced. For most papers the performance of the introduced method is better than the performances of earlier introduced methods. Some works however concentrate on e.g. lowering the number of parameters while maintaining a similar performance \cite{lust2020gran}.\\

\subsection{Findings}
\label{Sec:Combine}
\begin{figure*}[t]
	\centering
	\includegraphics[width=1.0\textwidth]{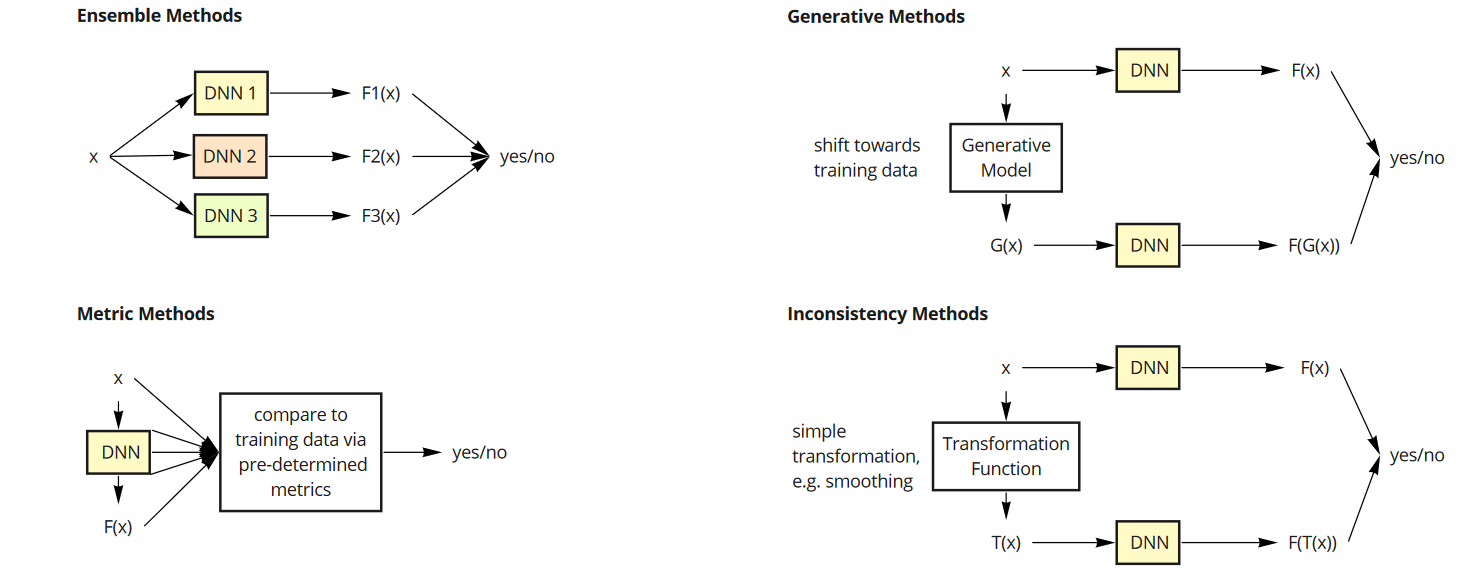}
	\caption{The structural bases of the methods of the four core principles: ensemble, generative, metric and inconsistency. At inference time they decide if the input sample $x$ is inside (yes) or outside (no) the generalization area and hence if the predicted class based on the output F(x) is expected to be correct.}
	\label{Fig:coreprinciples}
\end{figure*}
In Section \ref{Sec:Methods} we introduced methods to detect at inference time if an input is within the generalization area of a DNN. Currently, those methods can be found in three main literature fields: predictive uncertainty, out-of-distribution detection and adversarial example detection. Each literature field concentrates on a different reason for a sample to not be within the generalization envelope.\\
Methods in the \textbf{predictive uncertainty} field try to determine the probability of in-distribution samples to be misclassified. For this purpose they typically focus on samples that are close to a decision boundary and assign them a high uncertainty. Approaches based on this idea, however, are not capable of assigning a high uncertainty to samples that are not close to any decision boundary but still not within the generalization area.\\
Samples far away from the decision boundary and outside the generalization area are considered in the literature field \textbf{out-of-distribution detection}. A sample is out-of-distribution if it is different from the training samples in a principled manner.\\
The third literature field is \textbf{adversarial example detection}. Adversarial examples are constructed or selected such as to fool the DNN on purpose. For the construction of an adversarial image an in-distribution input image is slightly changed so that the relevant features for the DNN's decision point to the wrong class. The change in the image is hardly perceivable by humans or at least of a nature not impairing a correct decision. Adversarial examples are often neither close to a decision boundary in the features space, nor far away from in-distribution examples. This sets this field apart from predictive uncertainty and out-of-distribution.\\
These fields are currently considered separately, even though the same core principles are used to handle the underlying generalization deficiencies. These core principles lead to specific approaches, that we have identified and gathered into: ensemble methods, inconsistency methods, generative methods and metric methods. The structural base of each core principle is shown in Figure \ref{Fig:coreprinciples}.\\ 
\textbf{Ensemble methods} use several DNNs each trained slightly different. At inference time they are all applied on the input. The more their outputs differ the higher the chance that the decision for that input is wrong. This concept works well for samples that are close to a decision boundary, since for all the different networks the decision boundary varies slightly. However, out-of-distribution and certain types of adversarial samples are usually found far away from the decision boundary in the feature space. For those adversarial examples that are in the problem space or even close to the decision boundary, ensemble methods are less suited, as DNNs trained on the same data are vulnerable to the same attacks --- adversarial examples exhibit a closer relationship to the training dataset than to the training procedure itself \cite{ilyas2019adversarial}. Thus, most ensemble based methods are mainly found in the predictive uncertainty category and none in the field of adversarial example detection. The two methods in the out-of-distribution category are slightly different to those of the predictive-uncertainty since they use for training an additional data set containing samples from outside the expected generalization area (i.e., the problem space). During the training the DNNs are forced to produce strongly different outputs when receiving inputs from outside the expected generalization area.\\
\textbf{Inconsistency methods} use the idea, that for misclassified inputs the classification output is more sensitive to small changes in the input sample. A high difference between the outputs of an input image and the slightly transformed image indicates the input to be outside the generalization area. These assumptions fit particularly the adversarial examples detection setup and less the out-of-distribution and the predictive uncertainty setup. The inconsistency methods in the adversarial example field tend to fail for adversarial examples on more complex datasets since relatively simple noise-reduction procedures currently used were not able to remove the adversarial noise without distortion also important parts of the input \cite{xu2018feature}, \cite{ma2019nic}. The only inconsistency-based method for out-of-distribution images adds additional noise to the input such as to shift the sample away from the predicted class. This procedure shows some success but is outperformed by newer out-of-distribution detection methods using other core principles than inconsistency.\\
In the case of \textbf{generative methods} the images are shifted during a pre-processing procedure towards the generalization envelope, often assuming that this is found close to the training data. A difference between the outputs of the DNN computed from the shifted and the original input is then used as an indicator for an input outside the generalization area. In contrast to inconsistency methods here we do focus with the generator model on the setup that we know and for which we have evidence in the form of correctly classified samples. Intuitively, in-distribution samples are left untouched, while for the others, the generator model is supposed to eliminate the features that are responsible for this, which in turn is expected to lead to a shift in classification output. The underlying assumptions of generative methods are rather well suited for adversarial and out-of-distribution samples, but do not fit that well the predictive uncertainty setup.\\
The most promising methods are \textbf{metric methods}, which can be found in all three categories. They also hold the only method that is evaluated on an adversarial as well as on an out-of-distribution set-up \cite{lee2018simple}. They typically compare the DNN's outputs or gradients of several layers to the corresponding outputs of the training samples that have been investigated before using some metric. Unfortunately, they usually need an additional set with samples outside the generalization area for their training process in order to calibrate the threshold used for deciding when a sample lays outside the generalization area. However, as shown in some works \cite{Hein_2019_CVPR},\cite{lee2018simple}, it is possible to find methods that are able to train on one outlier class and to generalize well to a different one. Methods using this core principle often take a closer look on how the input is processed in the DNN. They take thus into consideration the possibility that the reasons for the lack of generalization might be hidden deep in the information flow within the DNN but are blurred in the actual output. Furthermore, the metric methods hold the most efficient methods both what the number of parameters and the computation time are considered, as described in Section \ref{Sec:Criteria}.\\
There are a lot of promising methods for each of the different literature fields, but due to the lack of evaluation variety for each method it is not possible to tell which method is best when applied to the combined task of detecting samples outside the generalization area. Most detection methods just focus on one of the literature fields. Only metric methods have been shown to achieve good results on more than one detection task (out-of-distribution and adversarial example detection).

\section{Conclusion}
\label{Sec:Conclusion}
There is a general interest for improving DNN performance by analyzing the generalization behavior. However, despite good performance in numerous problem setups, grasping the generalization behavior of DNNs represents an open research field. Its significance rises even more with the advent of safety-critical DNN applications like for example autonomous driving. In such cases understanding the generalization behavior represents a cornerstone of a coherent safety argumentation, which in turn is paramount for the wide public acceptance of such solutions.
\\ 
This paper presented a comprehensive survey covering the methods thought to detect at inference time if an input is within the generalization area of a DNN with a focus on the task of image classification.
\\
There can be different reasons for a sample not to be within the generalization area. In general we may either have to do with naturally occurring samples that lay outside the generalization area or in the case of adversarial examples, the input samples are selected on purpose or even engineered such as to lay outside the generalization area. However, the underlaying cause of the error in classification, irrespective of the setup, is that the DNN does not generalize correctly. From this perspective, all setups are equivalent and hence this paper reviews all corresponding contributions and sets them into the common broader generalization context.
\\
The generalization issues often stem from the fact that the feature projection learned during training is such that it contains bad modes that afflict separability. Thus, significant and decision-influencing differences between inputs do not correspond to large-enough distances in the feature space. So we either have no separability in the feature space or we have it but it is ignored during training.
\\
An analysis of the generalization behavior of a DNN can be used both during training and at inference time. During training the results of the analysis are used to build the optimal pair of feature extractor and decision surface given the available data. This setup has been studied extensively and it still lays in the focus of research.
\\ 
At inference time the common practice for investigating the generalization behavior, was to define confidence measures on the base of layer-wise neuron activations computed using the weights established during training. This however, leads often to misleading results, as the DNNs tend to be overconfident in their decision as measured by the relative significance allocated to the winning class. Predictive uncertainty methods improve upon this approach but usually focus mainly on the region of the input/feature space close to the decision surface. Methods of the "Out-of-distribution" literature field analyse the probability of the current input sample conditioned on the training data and may thus also look at regions far away from the decision surface. As adversarial examples may occur everywhere, methods to detect them come closest to a general approach for investigating the generalization area. However, they assume intent and accordingly have a tendency to ignore some blatant generalization issues, that do not involve looking at samples that are close in the input (image) space and far away in the feature space or the other way around.
\\       
A major approach in investigating for a certain input if the classification output can be trusted, irrespective of the literature field is to analyse if the input is within the range of the training data. In particular for metric methods, and to a lesser extent for inconsistency methods, this approach is complemented by one targeting the way information is processed within the DNN.
\\  
Given the fact that until now the generalization questions have been treated mostly in separated rather application-related setups, it is difficult to tell beyond doubt which method or class of methods holds the best promise for deciding if a sample is within the generalization area of a DNN. In general there are many promising approaches. For example, metric methods are currently in the focus of research in all literature fields when judging by the number of publications. In particular the adversarial-example setup constitutes a very active research field that we believe has much to profit from becoming more aware of the other related fields.\\ 
{\small
\bibliographystyle{ieee_fullname}
\bibliography{generalization_survey}
}

\end{document}